# Detecting AI-Generated Text in Educational Content: Leveraging Machine Learning and Explainable AI for Academic Integrity

Ayat A. Najjar
Faculty of Modern Science
Arab American University
13 Zababdeh, P.O Box 240 Jenin, Palestine
ayat.najar@aaup.edu

Huthaifa I. Ashqar*
Civil Engineering Department
Arab American University
13 Zababdeh, P.O Box 240 Jenin, Palestine
Artificial Intelligence Program
Fu Foundation School of Engineering and Applied Science
Columbia University
500 W 120th St, New York, NY 10027, United States
huthaifa.ashqar@aaup.edu

Omar A. Darwish
Information Security and Applied Computing
Eastern Michigan University
900 Oakwood St, Ypsilanti, MI 48197, United States
odarwish@emich.edu

Eman Hammad
iSTAR Lab
Engineering Technology & Industrial Distribution
Texas A&M University
400 Bizzell St, College Station, TX 77840, United States
eman.hammad@tamu.edu

*Corresponding Author

# Detecting AI-Generated Text in Educational Content: Leveraging Machine Learning and Explainable AI for Academic Integrity

**Abstract**

This study seeks to enhance academic integrity by providing tools to detect AI-generated content in student work using advanced technologies. The findings promote transparency and accountability, helping educators maintain ethical standards and supporting the responsible integration of AI in education. A key contribution of this work is the generation of the CyberHumanAI dataset, which has 1000 observations, 500 of which are written by humans and the other 500 produced by ChatGPT. We evaluate various machine learning (ML) and deep learning (DL) algorithms on the CyberHumanAI dataset comparing human-written and AI-generated content from Large Language Models (LLMs) (i.e., ChatGPT). Results demonstrate that traditional ML algorithms, specifically XGBoost and Random Forest, achieve high performance (83% and 81% accuracies respectively). Results also show that classifying shorter content seems to be more challenging than classifying longer content. Further, using Explainable Artificial Intelligence (XAI) we identify discriminative features influencing the ML model's predictions, where human-written content tends to use a practical language (e.g., use and allow). Meanwhile AI-generated text is characterized by more abstract and formal terms (e.g., realm and employ). Finally, a comparative analysis with GPTZero show that our narrowly focused, simple, and fine-tuned model can outperform generalized systems like GPTZero. The proposed model achieved approximately 77.5% accuracy compared to GPTZero's 48.5% accuracy when tasked to classify Pure AI, Pure Human, and mixed class. GPTZero showed a tendency to classify challenging and small-content cases as either mixed or unrecognized while our proposed model showed a more balanced performance across the three classes.

## 1. Introduction

Our communication practices are quickly changing due to the emergence of generative AI models. It is widely used in various disciplines, including healthcare, academic research, the arts, and content production. Large Language Models (LLMs) has demonstrated performance in comprehending user inquiries and producing text that resembles human speech. LLMs attracted wide attention for researchers, policymakers, and educators. Although LLMs is claimed to have the ability to transform society, there are some potential risks as well. The advent of innovative AI-based chatbots, emphasizes the need to questions the originality of the ideas, languages, and solutions (i.e., whether a sentence was generated by an AI or by a human). Investigating the originality of a written idea has significant effects in several sectors including digital forensics and information security. Defensive measures are necessary to prevent increasingly complex attacks that exploit textual content as a potent weapon due to the dynamic nature of cybersecurity, such as the transmission of false information and disinformation or social engineering attempts [1]. Specifically, in the field of information security, where the ability to recognize AI-generated material is vital, a detrimental application of AI is required. Additionally, it can spread false information and fake news throughout online platforms [2]. Moreover, LLMs may also provide inaccurate answers and information since they were trained on outdated data, or they may suffer from hallucination [1], [3].

As the use of generative AI tools like ChatGPT becomes widespread in education [4], [5], [6], ensuring that student work is genuinely human-authored is a growing concern for educators. By developing a robust model to detect AI-generated text, this study provides a valuable tool for educators, promoting fairness and academic integrity. The broader educational relevance lies in its application to plagiarism detection, the evaluation of digital submissions, and the safeguarding of learning outcomes in environments where AI-generated content is increasingly utilized. This contributes to the field's understanding of how digital technologies can support both pedagogy and ethical standards in education.

Academic institutions additionally highlight the issue of plagiarism for students who utilize these tools to produce their homework and term papers. Most academic courses require strong writing abilities. Having students who rely solely on tools like LLMs would result in a generation of students lacking the ability to express themselves properly [2]. Writing skills are a key part of most academic courses. It becomes essential to develop strategies for recognizing plagiarism and to verify the quality and dependability of information. This study aims to develop a simple and explainable Machine Learning (ML) model that can recognize cybersecurity written documents generated by LLMs. This will be accomplished by creating a new dataset and use it to investigate detecting human-written and AI-generated text. We will also develop classic ML classification algorithms, Deep Learning (DL), and Explainable AI (XAI). The main contributions of this study can be summarized as follows:

1. Generates the CyberHumanAI dataset, which includes 1000 observations of ChatGPT/Human cybersecurity paragraphs.
2. Advances knowledge on the pedagogical use of digital technology to improve learning environments and maintaining educational standards.
3. Shows the capability of XGBoost and Random Forest as traditional Machine Learning models to classify AI-generated content with high accuracy and minimum misclassification rates.
4. Shows that classifying relatively smaller content (e.g., paragraphs) is more challenging than classifying larger ones (e.g., articles).
5. Utilizes XAI to provide transparency and explainability for the classification results, which identifies key features that differentiate human and AI-generated content.
6. Compares GPTZero and our proposed model, which shows that a narrow, simple, fine-tuned AI system can outperform a generalized AI system like GPTZero in our specific task.

## 2. Related Work

The growing use of generative AI models in the arts, academics, healthcare, and content creation is a fast-changing communication method [7], [8], [9]. The difference between text produced by humans and by AI must be highlighted to identify and address the potential influence in these fields. These two studies [10], [11] worked on the difference between AI and human text generated. The first study [10] presented a paradigm for recognizing AI-generated material, especially in academic and scientific writing. A model is trained using predetermined datasets, and it is then deployed on a cloud-based service. The suggested framework, which made use of artificial neural networks, obtained an accuracy of 89.95% compared to tools like OpenAI Text Classifier (42.08%) and ZeroGPT (87.5%). The other study [11] examined a plausible situation in which text is converted from human-written to AI-generated texts using neural language models. It showed that annotators have difficulty with this activity but can become better with rewards. To encourage further research in human text recognition and evaluation, the study analyzed several aspects influencing human detection performance, including model size and prompt genre. It also introduces the RoFT dataset with 21,000 human annotations and mistake classifications [11].

Academic institutions raised concerns about plagiarism because more students are using LLMs for their assignments and term papers, which could affect their writing abilities. Many studies concentrated on recognizing LLM-generated content across different areas such as [12], [13], [14], [15], [16], [17], [18], [19]. This study [12] examined the distinctions between medical writings produced by ChatGPT and those authored by human specialists. It also developed ML processes to identify and distinguish between ChatGPT-produced medical texts efficiently. They tested various models, including perplexity-CLS, CART, XGBoost, and BERT. They discovered that the F1 of the BERT-based model reached 95%, making it capable of accurately detecting medical texts produced by ChatGPT. Another study [13] compared human-written and ChatGPT-generated text in two experiments to examine the effectiveness of short online reviews. The first experiment used ChatGPT content produced by bespoke queries, and the second uses text produced by paraphrasing initial human-generated reviews. They discovered that when employing rephrased language, the ML model had a harder time distinguishing between human and ChatGPT-generated reviews than when using a perplexity score-based strategy. However, the accuracy of their suggested method was reached 79%. In addition, another study [14] compared eleven classification methods to distinguish between text produced by ChatGPT and text written by humans. The suggested model obtained about 77% accuracy when applied to GPT-3.5 generated text in tests on a Kaggle dataset of 10,000 documents, comprising of about 5,204 human-written texts from news and social media. Another study [15] introduced a dataset, CHEAT, to aid in the development of detection algorithms and first investigate the potentially detrimental effects of ChatGPT in academia. It contains 35,304 synthetic abstracts, with the primary varieties being Polish, Generation, and Mix. ChatGPT-written abstracts were detectable, according to an analysis of text synthesis detection techniques, with detection difficulty rising as human involvement does as well.

In the study [16], the authors investigated techniques for differentiating AI-generated from rephrased text, including instances when AI imitates human writing. They employed a diverse text corpus and produced a relatively high F1 score, exceeding 96% for both simple and complex human/AI-generated language, and over 78% for rephrased material. It is noteworthy that their top basic text rephrasing detection algorithm exceeded GPTZero's F1-score. The TSA-LSTMRNN model, which combines the Tunicate Swarm Algorithm and Long Short-Term Memory Recurrent Neural Network, was introduced in [17] and used to identify text that was produced by human and ChatGPT. It extracted features using TF-IDF, word embedding, and count vectorizers, classifies them using LSTMRNN, and optimized parameters using TSA. On benchmark datasets, the results demonstrated performance with a maximum accuracy of about 93.2% for text created by humans and about 93.8% for text generated by ChatGPT.

Through a comparison of expertly written and under-reviewed articles, as graded by two musculoskeletal radiologists [18], a study assessed the correctness of ChatGPT-generated radiology articles. It was discovered that four out of five articles produced by ChatGPT were highly wrong and contained false references. The introduction and discussion of one piece were well-structured, but it also had wholly made-up references. In [19], authored evaluated ML algorithms to distinguish between human and AI-generated text. Random Forests (RF) notably achieved the highest accuracy of about 93%, highlighting its potential in content moderation and plagiarism detection.

However, our study aims to differentiate between text written by humans and text generated by LLMs using ML and DL. This study is a groundbreaking endeavor as it will generate dataset of articles in the field of cybersecurity. To better understand the behavior of our model, we used XAI, emphasizing transparency and interpretability. We recognize the importance of understanding not only the model's outputs but also its underlying processes. We also conduct a thorough comparison between our model's accuracy and that of GPTZero, a well-known commercial AI product, as an essential benchmark.

## 3. Methodology

Generally, data preparation and feature selection processes from the generated dataset play important roles in simplifying the overall subsequent tasks, like the classification task, and therefore leading to improved classification rates. This study proposes a framework for detecting the cybersecurity documents written by ChatGPT, shown in Figure (1), which includes five main phases including data preprocessing, feature selection, ML model, document detection and classification, and explainable AI. The following sections describe each step of this framework.

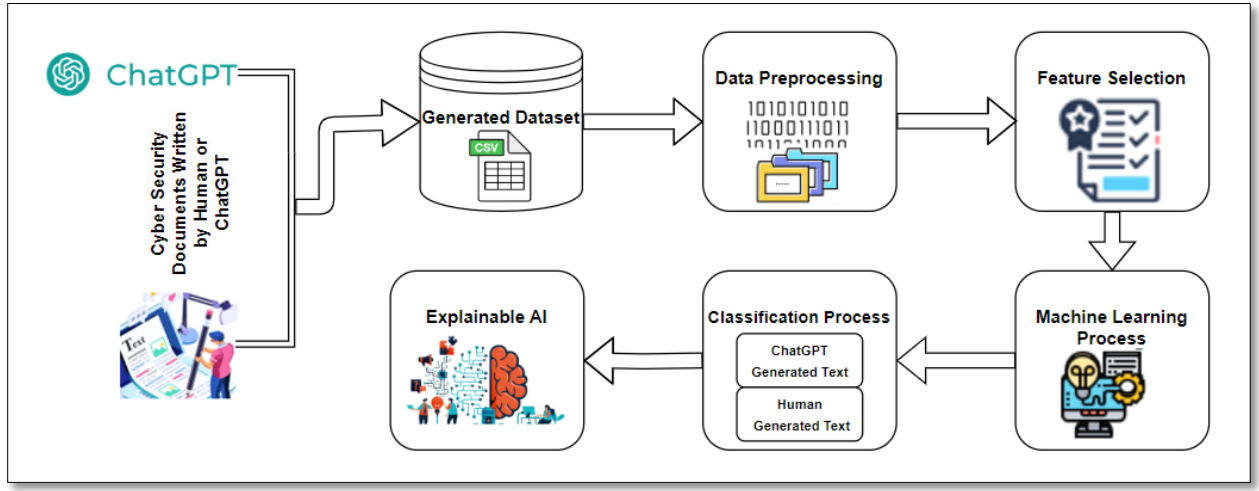

**Figure (1):** The general workflow of the proposed ML model for detecting and classifying the cybersecurity documents written by ChatGPT.

*3.1 Dataset*

We generated ChatGPT/Human cybersecurity paragraph dataset, which has about 1000 observations and was compiled in September 2023. It has 500 paragraphs written by humans and another 500 produced by ChatGPT, all of which are on cybersecurity and share the same title. This dataset to acts as a fundamental step for creating machine-learning models capable of differentiating ChatGPT-generated cybersecurity documents. The human-written cybersecurity paragraphs were extracted from Wikipedia API using Python and through the keyword *computer security*. This unique generated dataset offers a great tool for researchers and practitioners who want to investigate and address cybersecurity document categorization problems using ML methods.

A preliminary check was done to find and remove empty observations. We prepared the dataset by stop words removal, lemmatization, punctuation removal, and tokenization of the text [20] putting the text data into a clean, structured format suited for classification and model creation. The word cloud for the two classes human and ChatGPT is displayed in Figure (2) (a) and Figure (2) (b), respectively. Table (1) displays the word frequency for the two classes as counts and percentages. Results compares the frequency of words used by humans and ChatGPT, which highlight the differences in their vocabulary when discussing topics related to security and computing. The word "security" is the most frequent for both, with humans using it 420 times (1.71%) and ChatGPT using it 411 times (1.52%). However, differences emerge with other terms: humans tend to use "use" (312 counts, 1.27%) more frequently, while ChatGPT emphasizes "system" (261 counts, 0.97%) and "computer" (233 counts, 0.86%) more than humans. Notably, "information" is used more often by humans (206 counts, 0.84%) compared to ChatGPT (166 counts, 0.61%). Humans and ChatGPT show similar trends with some variation in the emphasis of technical terms.

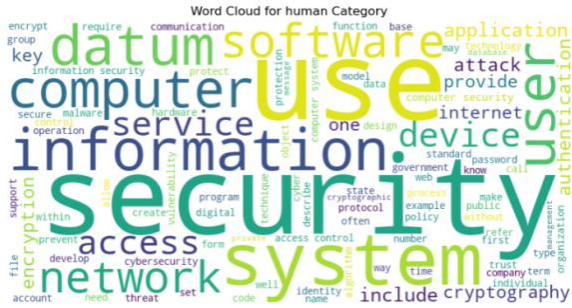 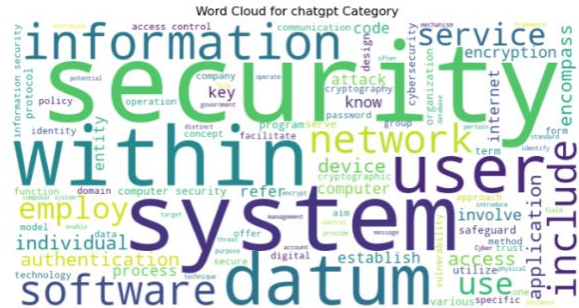

(a) Word cloud for 'human' class  (b) Word cloud for 'chatgpt' class

Figure (2): Word cloud for 'human' and 'chatgpt 'classes.

Table (1): Top 10 words for 'chatgpt' and 'human' classes (counts and percentage of total tokens)

| Words Frequency – Human | | | Words Frequency – ChatGPT | | |
| --- | --- | --- | --- | --- | --- |
| Word | Counts | Percentage % | Word | Counts | Percentage % |
| security | 420 | 1.71 | security | 411 | 1.52 |
| use | 312 | 1.27 | system | 261 | 0.97 |
| system | 264 | 1.07 | computer | 233 | 0.86 |
| computer | 251 | 1.02 | within | 220 | 0.81 |
| information | 206 | 0.84 | datum | 183 | 0.68 |
| datum | 160 | 0.65 | information | 166 | 0.61 |
| user | 158 | 0.64 | access | 153 | 0.57 |
| access | 155 | 0.63 | user | 149 | 0.55 |
| software | 115 | 0.47 | authentication | 114 | 0.42 |
| network | 114 | 0.46 | software | 113 | 0.42 |

The dataset was then separated into training and testing subsets, with 80% of the data going toward training and 20% going for testing. This division played a critical role in the model evaluation process by evaluating the model's performance on unseen data. Subsequently, a TF-IDF (Term Frequency-Inverse Document Frequency) Vectorizer was used to make it easier to convert the text data into a machine-learning-friendly format [21]. By transforming the text data into a matrix of numerical features, this approach was able to capture the significance of words inside each document while considering their frequency across the entire dataset. With "0" denoting the ChatGPT class and "1" denoting the Human class, the resulting TF-IDF vectors served as the basis for training ML models on this dataset, allowing the creation of classifiers to differentiate between human and ChatGPT-generated cybersecurity paragraphs. Figure (3) displays the top 10 words with the highest TF-IDF weights for the two classes, "human" and "chatgpt". In both cases, "security" stands out with the highest weight, approximately 23 for humans and about 22 for ChatGPT, which further highlights its significant importance in both sets. However, the other terms show notable differences. In human text, words like "use" and "computer" have higher weights, around 16 and 14, respectively. In ChatGPT text, words such as "datum" and "authentication" have more emphasis, each close to 11. Moreover, terms like "employ," "realm," and "encryption" appear only in ChatGPT's word list, which shows a distinct vocabulary focus compared to human-written text. The differences show ChatGPT's preference for more technical and system-related terms, while humans emphasize broader concepts like "use" and "computer."

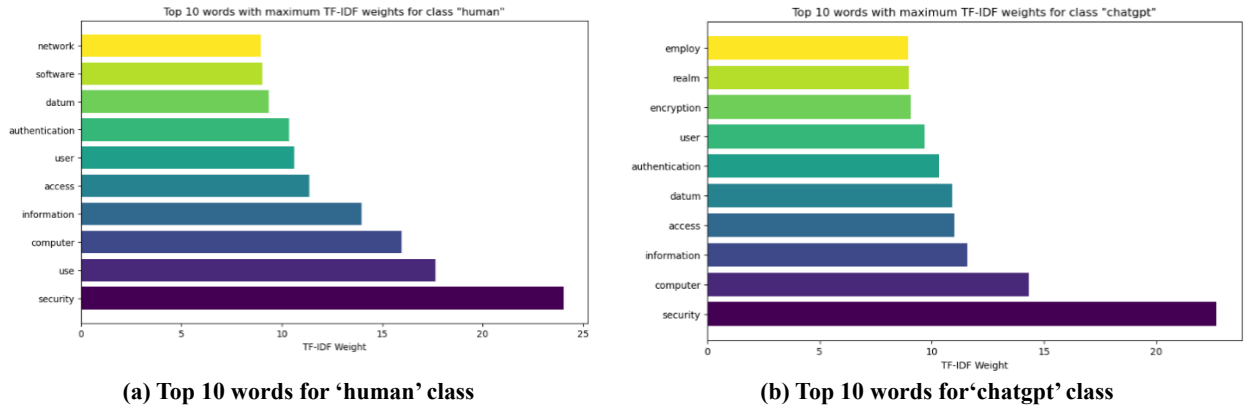

(a) Top 10 words for 'human' class          (b) Top 10 words for 'chatgpt' class

**Figure (3): Top 10 words with maximum TF-IDF weights for 'human' and 'chatgpt' classes.**

### *3.2 Classification Algorithms*

The process of training models to generate predictions and categorize the cybersecurity documents written by ChatGPT was done using a variety of algorithms. This method enables us to compare between different algorithms that can recognize patterns in data and take actions based on those patterns. We used a variety of ML techniques, including RF, Support Vector Machines (SVM), J48, and XGBoost. Each carefully crafted to quickly explore and categorize cybersecurity content to detect plagiarism [22] .The following is a brief description of the used algorithms.

RF is a robust ensemble learning method and it is well-known for its performance in both classification and regression applications. With the help of a group of decision trees, it performs well at managing complex datasets and reducing overfitting. Random Forest is a highly favored option in numerous fields due to its adaptability and resilience, yielding precise and dependable forecasts as well as valuable insights via feature importance analysis [23], [24], [25].

SVM is a strong supervised learning algorithm that may be used for regression and classification tasks. SVM is known to perform well if data is divided into different classes and maximizing the margin between them by finding the best hyperplanes. Known for their adaptability and efficiency in high-dimensional areas, SVMs are an important ML tool for producing precise and dependable predictions [26], [27], [28].

J48 is known for its ease of use and interpretability in ML as a common decision tree classifier. J48 recursively divides data according to attribute values, generating a tree structure for effective decision-making, and is based on the C4.5 algorithm. J48 is a highly valuable tool in data mining and classification jobs due to its reputation for handling both numerical and categorical data. It offers clear and actionable insights for well-informed decision support [29], [30].

XGBoost is known as Extreme Gradient Boosting, which is an ML technique that is notable for its great efficiency and scalability. The XGBoost algorithm performs well in predictive modeling applications. XGBoost is a preferred option in many industries, including finance and healthcare, due to its capacity to manage complicated relationships in data, regularization techniques, and parallel processing. Acknowledged for its swiftness and efficiency, XGBoost has established itself as a mainstay in both practical and competitive ML scenarios [31], [32].

The study also explored the rapidly changing neural network landscape, utilizing the power of Convolutional Neural Networks (CNN) and deep Neural Networks (DNN) with a focus on cybersecurity document analysis [33]. These cutting-edge deep learning approaches produced ground-breaking capabilities, automating the complex pattern extraction from cybersecurity literature.

CNNs are an essential part of deep learning and are specifically engineered for computer vision and image recognition applications. CNNs, which consist of layers for convolution, pooling, and fully connected operations, are particularly good at recognizing complex spatial hierarchies in image data. CNNs are extremely useful in many different applications, such as object detection and picture classification, because of their capacity to automatically learn features from unprocessed data. CNNs are widely used in the field and are still driving discoveries in artificial intelligence, particularly in the areas of pattern recognition and visual perception [34], [35].

DNNs serve as a cutting-edge type of multi-layered ML models that can extract complex patterns and representations from data. DNNs have demonstrated their usefulness in various fields, such as image analysis, audio recognition, and natural language processing, by utilizing advanced architectures including feedforward and recurrent structures. Because of their ability to learn and abstract features hierarchically, DNNs are an effective tool for tackling challenging issues and opening new possibilities for a variety of industries [36], [37], [38].

The study aims to improve the detection and classification of cybersecurity papers produced by ChatGPT by integrating classical ML with cutting-edge DL techniques. The study uses different the performance metrics for ML models [39]such as accuracy, F1 score, precision, recall, confusion matrix, and ROC curve (Receiver Operating Characteristic Curve) in the search for a comprehensive solution, further strengthening its contribution to the protection of digital security in the contemporary threat scenario.

*3.3 Explainable Artificial Intelligence (XAI)*

In recent years, artificial intelligence has advanced significantly, sparking interest in previously understudied fields. The focus has shifted from solely focusing on model performance as AI advances to requiring experts to look at algorithmic decision-making processes and the logic behind AI models' output. As modern ML algorithms, especially deep learning, using black box techniques become more powerful and complex. It becomes more difficult to understand how they behave and why specific outcomes were achieved, or mistakes were made. XAI systems can be used to understand models' behaviors, which allow users to develop the proper level of trust and reliance [40], [41].

In this study, we use Local Interpretable Model-agnostic Explanations (LIME) to provide a way to understand how ML model make decisions. Because LIME is based on a model-agnostic premise, which was developed by Ribeiro et al. in 2016 [42], it can offer visible and interpretable insights into the predictions of different black-box models. LIME generates locally faithful approximations through perturbed samples around individual instances, enabling users to understand the reasoning behind individual predictions. Its interpretability-enhancing capabilities and adaptability have led to LIME's widespread adoption in various domains, where it is a valuable resource for researchers and practitioners seeking transparency in the decision-making process of complex ML algorithms.

## 4. Experimental Results

This section assesses the performance of different ML algorithms using an 11th generation Intel(R) Core (TM) i5-1135G7 @ 2.40GHz processor, 16.0 GB of RAM, and a 64-bit operating system. We started our experiment by investigating whether the use of full articles or paragraphs as the main unit of comparison produces different results. The comparison is shown in Table (2) and Table (3), which highlights the performance of different algorithms in distinguishing cybersecurity articles and paragraphs generated by ChatGPT. In Table (2), algorithms such as J48 and XGBoost achieve the highest accuracy, precision, recall, and F1-score of 100% in identifying ChatGPT-generated articles, followed closely by RF, DNN, and CNN, which all have a high accuracy of 99%. In contrast, Table (3) shows lower accuracy across all algorithms for distinguishing ChatGPT-generated paragraphs. XGBoost performs the best with an accuracy of 83%, while RF follows closely at 81%. Other models like SVM and DNN show a relatively lower accuracy, with DNN reaching only 69%. The difference in performance between article and paragraph detection shows that distinguishing between smaller text segments (paragraphs) is more challenging for the models compared to longer, more structured articles.

**Table (2): Accuracy for distinguishing the cybersecurity articles generated by ChatGPT.**

| Algorithm | Accuracy | Precision | Recall | F1-Score |
|-----------|----------|-----------|--------|----------|
| RF | 99.0% | 99.0% | 99.0% | 99.0% |
| SVM | 97.0% | 97.0% | 96.0% | 96.0% |
| J48 | 100% | 100% | 100% | 100% |
| XGBoost | 100% | 100% | 100% | 100% |
| DNN | 99.0% | 99.0% | 99.0% | 99.0% |
| CNN | 99.0% | 99.0% | 99.0% | 99.0% |

**Table (3): Accuracy for distinguishing the cybersecurity paragraphs generated by ChatGPT.**

| Algorithm | Accuracy | Precision | Recall | F1-Score |
|-----------|----------|-----------|--------|----------|
| RF | 81.0% | 82.0% | 82.0% | 81.0% |
| SVM | 70.0% | 70.0% | 70.0% | 70.0% |
| J48 | 72.0% | 72.0% | 72.0% | 72.0% |
| XGBoost | 83.0% | 84.0% | 84.0% | 83.0% |
| DNN | 69.0% | 70.0% | 70.0% | 69.0% |
| CNN | 79.0% | 79.0% | 79.0% | 79.0% |

The results show that ML algorithms were generally better at distinguishing cybersecurity articles generated by ChatGPT than cybersecurity paragraphs generated by ChatGPT. In one hand, ChatGPT articles seems to only include text that significantly differed from the original human-written articles from Wikipedia. The latter seemed to be retrieved completely, which included links, symbols, and other non-alphabetic features. Additionally, the results demonstrate that deep learning algorithms are not as effective as standard ML methods. This could be due to several factors. For example, deep learning algorithms work best with larger datasets, while classical ML methods could be better suited for smaller datasets.

*4.1 Machine Learning and Deep Learning Results*

As the results from using paragraphs instead of articles showed a more challenging problem, this section and the later ones will focus on these results. This section discusses the results by investigating the confusion matrix for four different ML methods (i.e., RF, SVM, XGBoost, and J48) and two DL algorithms (i.e., DNN and CNN) as shown in Figure (4). The confusion matrices in Figure (4) provide insights into the performance of various algorithms in differentiating between human-written and ChatGPT-generated content. XGBoost, shown in Figure (4) (c), demonstrated relatively the highest performance. XGBoost was able to classify 42.42% of ChatGPT-generated content and 40.91% of human content, with minimal misclassification of 11.11% and 5.56%, respectively. For RF in Figure (4) (a), it follows closely with about similar results, correctly identifying 40.91% of ChatGPT-generated and 40.40% of human-written content. Nonetheless, SVM and J48 in Figure (4) (b) and (d) showed slightly higher misclassification rates for human-generated content, with SVM incorrectly labeling about 11.11% and J48 misclassifying about

15.15%. In the case of the DL results, the two algorithms have lower accuracy for ChatGPT detection, with DNN only identifying about 32.83% of ChatGPT content correctly, while CNN performs slightly better at 41.41%. XGBoost and RF outperform the others, showing higher accuracy in distinguishing between the two types of content.

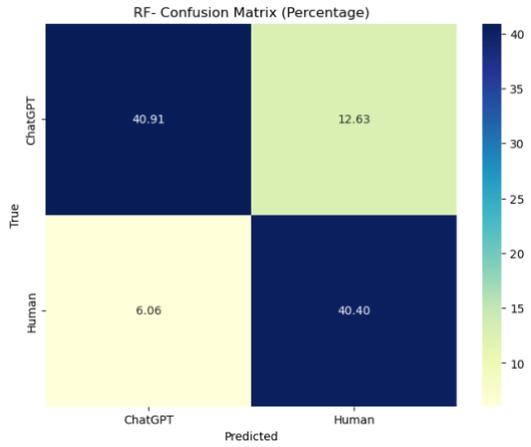
(a) RF

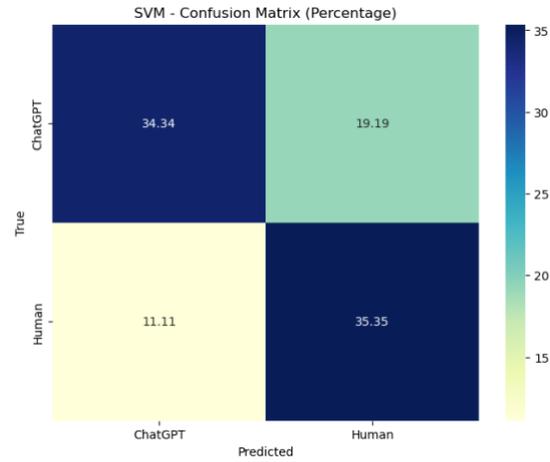
(b) SVM

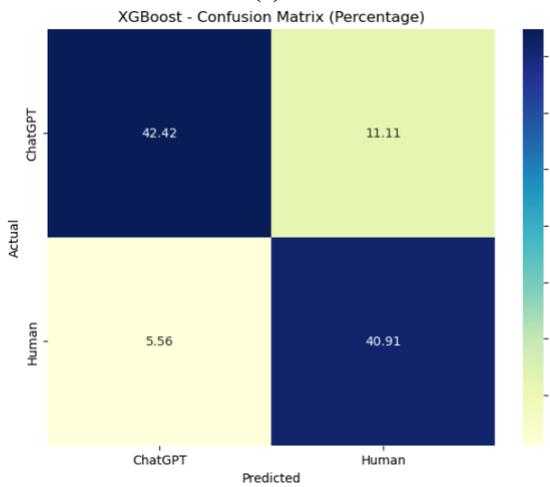
(c) XGBoost

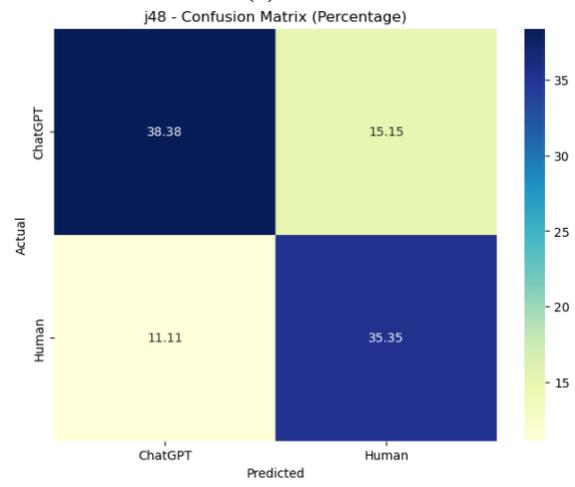
(d) J48

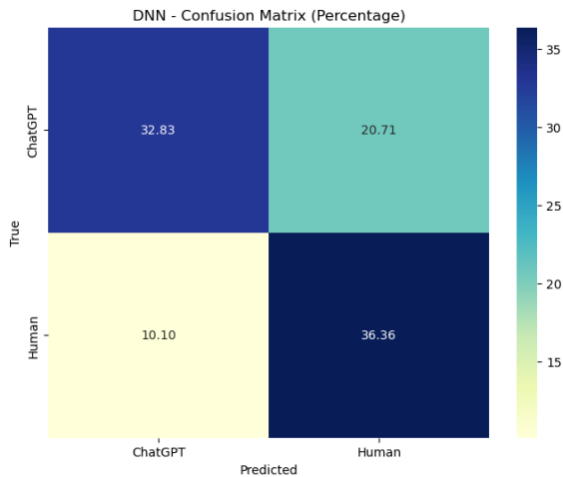
(e) DNN

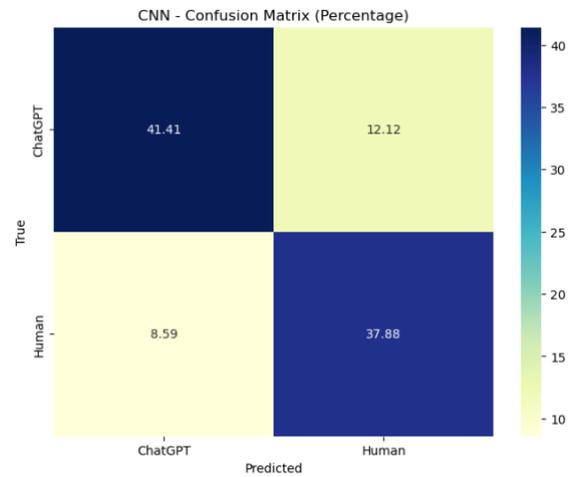
(f) CNN

**Figure (4): Confusion matrices for the six used algorithms.**

The results show the ability of ML algorithms to distinguish between human-written and ChatGPT-generated cybersecurity content with high accuracy. The highest-performing algorithms, such as XGBoost and RF, showed minimal misclassification rates, which indicates that systems built on these algorithms can effectively discriminate AI-generated text from human-authored content. This is significant in the context of cybersecurity, where distinguishing between human-authored reports and AI-generated text could be critical for ensuring the integrity and trustworthiness of information. Automated systems could flag AI-generated phishing emails, preventing malicious content from being passed as genuine human communication. These findings have various applications. In academic and professional writing, this distinction can help identify plagiarism or ensure that content labeled as human-generated is truly authored by a person, maintaining ethical standards and providing reliability among stakeholders. In content moderation, platforms can use such algorithms to filter out AI-generated misinformation or disinformation, especially in sensitive areas like politics, finance, and news. Moreover, businesses using ChatGPT for customer service or automated report generation could ensure that human oversight is applied to verify critical information. This can save the time but improving the reliability and accountability of their operations.

## *4.2 Explainable AI (XAI) Results*

As XGBoost algorithm achieved the highest performance in detecting AI-generated cybersecurity text, we employed LIME as an XAI to deeply explain the classification results of XGBoost. By providing insights into the reasons impacting the model's conclusions in the field of cybersecurity, this technique improves transparency and reliability. Figure (6) shows the top ten important features (i.e., words) for the human and ChatGPT classes in the XGBoost model, as generated by LIME. It clarifies the precise words that have a major influence on the model's predictions in each class. Figure (5) shows the interpretability of LIME on the local level. This helps to clarify the decision-making process of the black-box model by providing real insights into the critical characteristics driving classification results for various text categories. For the human class, terms like "allow," "use," "virus," and "people" are considered highly discriminative, which indicates that humans tend to use more practical, action-oriented language related to security (e.g., viruses, prevention, and business terms). However, the ChatGPT class is dominated by more of an abstract and more formal words such as "realm," "employ," "serve," and "establish," which reflects a more structured, generalized tone common in AI-generated content.

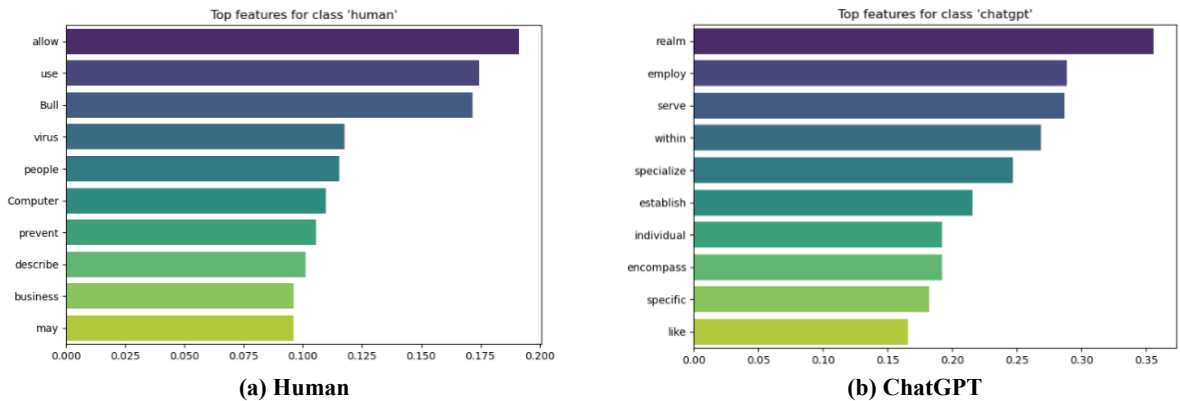

(a) Human  (b) ChatGPT

**Figure (5): The top ten important features for the "human" and "chatgpt" classes in the XGBoost model.**

To take this investigation deeper to the observation level, Figures (6) and Figure (7) explains the discriminative features on an instance (i.e., a paragraph) using LIME. The analysis in the two figures shows the local decision-making process of the model for an instance text. It shows the true label, predicted label, and a visual representation of the top ten features impacting the prediction locally. The instance that we used as an example is:

*Intel Software Guard Extensions SGX collection instruction code integrate specific Intel central processing unit cpu establish trust execution environment instruction enable userlevel operating system code establish secure private memory region call enclave SGX design application secure remote computation protect web browse digital right management DRM mind also find utility conceal proprietary algorithm encryption key SGX mechanism involve cpu encrypt section memory know enclave Data code originate within enclave decrypt realtime within cpu prevent inspection access code include code operate high privilege level like operating system underlie hypervisor although approach mitigate many form attack do not safeguard sidechannel attack shift Intels strategy 2021 lead removal SGX 11th 12th generation Intel Core Processors development SGX continue Intel Xeon processor intend cloud enterprise application.*

This paragraph was generated using ChatGPT and thus the true label is "chatgpt". The model was able to correctly predict the class and thus the predicted label is also "chatgpt". It also can be observed that the model predicted the class with about 99% accuracy. We can also see that the top three discriminative features are "safeguard," "establish," and "specific." The last two features were also highlighted in Figure (5) among the top ten discriminative features by the XGBoost algorithm used in this study.

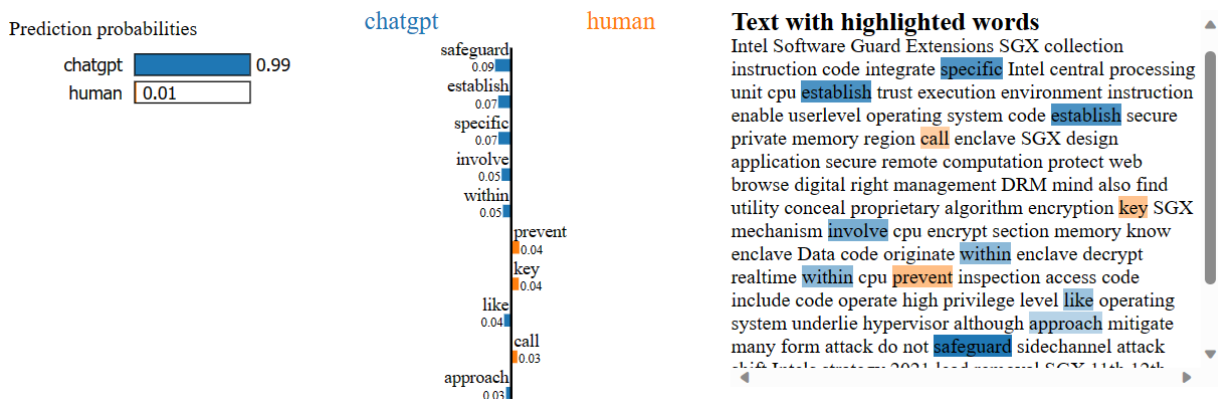

**Figure (6): The prediction probabilities using LIME for an instance in the data.**

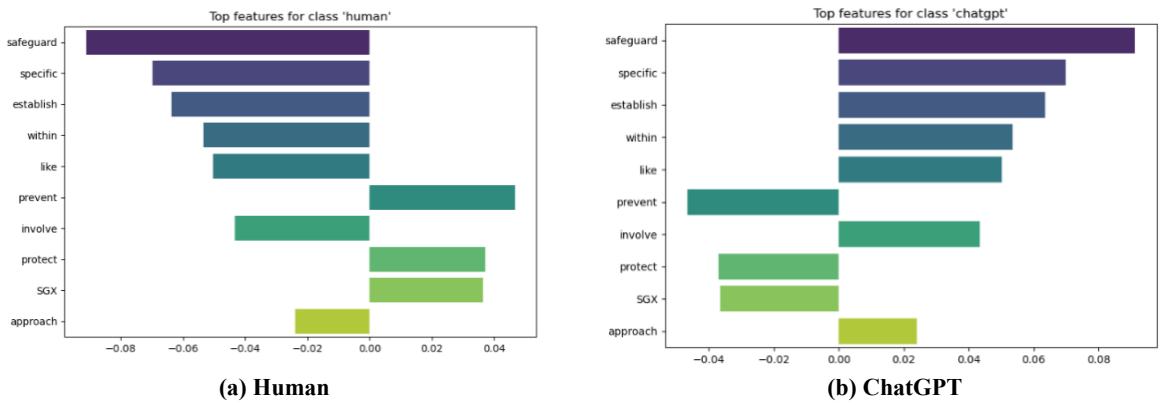

(a) Human  (b) ChatGPT

**Figure (7): The top ten important features for an instance in the data.**

*4.3 Comparison with GPTZero*

In this section, we compared our model's accuracy to a widely used software developed by the industry. The main goal is to benchmark our highest-performed model (i.e., XGBoost) against GPTZero [43], [44]. Doing so, we are not only verifying our model and its potential to be further scaled and transferred but also providing insights about the performance of GPTZero.

GPTZero, which was introduced in 2023 to address worries about AI-driven academic plagiarism, has received praise for its work but has also drawn criticism for producing false positives, particularly in situations where academic integrity is at risk [43], [44]. The program uses burstiness and perplexity metrics to identify passages that are created by bots [45]. Burstiness examines phrase patterns for differences, whereas Perplexity measures text randomization and odd construction based on language model prevalence. Human text has greater diversity than content generated by AI. In previous studies comparing GPTZero and ChatGPT's efficacy in assessing fake queries and medical articles [46], GPTZero was utilized. The study found that GPTZero had low false-positive and high false-negative rates. Another study of more than a million tweets and academic papers looked at opinions regarding ChatGPT's capacity for plagiarism [47]. In this study, we will investigate GPTZero for cybersecurity texts.

From the generated dataset of this study, we created new observations for this task. We divided 600 observations into three classes by creating combinations of text generated by ChatGPT and humans, as shown in Table (4). The first class includes only AI-generated text, labeled as Pure AI class. The second class includes a mix of human- and AI-generated texts of different ratios. The third class includes only human-written text. This split reflects the reality as ChatGPT was documented to be used in the two different forms; mixed with human-written, which sometimes referred to as paraphrased, and pure ChatGPT-generated text. We used 400 observations as training dataset for our model and 200 observations as a testing dataset.

**Table (4): Accuracy for distinguishing the cybersecurity paragraphs generated by ChatGPT.**

| Class Name | Number of Observations | ChatGPT Text Percentage |
|---|---|---|
| Pure AI | 200 | 100% |
| Mixed | 200 | 1%-99% |
| Pure Human | 200 | 0% |

The performance of GPTZero is shown in Table (5) and Table (6). Out of 200 observations, GPTZero was not able to classify 32 observations. The model showed an accuracy of 48.5% after adjusting for the 32 unrecognized cases, which are not factored in the confusion matrix shown in Table (6). GPTZero seems to perform exceptionally well in classifying mixed cases, with about 76 correct predictions and no misclassifications from the testing dataset. Nonetheless, it seems to struggle in identifying Pure AI and Pure Human instances. Only 3 Pure AI and 18 Pure Human instances were correctly classified, while 56 and 15 were misclassified as mixed, respectively. This suggest that the model is overly cautious or unable to distinguish clear patterns between human and AI content in many cases.

On the other hand, the performance of our proposed model using XGBoost is shown in Table (5) and Table (7). Results showed a more balanced classification performance across all classes, with an accuracy of about 77.5% and no unrecognized cases. The model was able to identify about 48 out of 66 of the Pure AI instances and performs relatively well on the mixed and Pure Human classifications, with 55 and 52 correct predictions, respectively. Misclassifications are still present but seems to be lower than those resulted from GPTZero, especially for mixed and Pure Human instances. The model was also able to better identify Pure Human cases compared to GPTZero, with 52 out of 67 instances correctly classified.

The differences between GPTZero and our proposed model can be explained by their design goals and training data. GPTZero seems to be likely designed to be more cautious and conservative. It tends to classify uncertain cases as either mixed or unrecognized rather than taking the risk of misclassifying them as Pure AI or Pure Human. This results in high precision for the mixed cases but a lower performance for the other classes. The other reason is for this disparity is that GPTZero had trouble identifying text that had less than 250 characters [48]. Nonetheless, our proposed model shows a more balanced performance, with fewer misclassifying cases as mixed. This indicates that our proposed model was trained on a more specific dataset, which made it more fine-tuned to better capture the discriminative features between AI-generated and human-written content. This suggests that using a narrow AI system fine-tuned with a suitable dataset in a specific task can beat a more generalized AI systems.

**Table (5): Comparison with the GPTZero tool.**

| Class Name | Accuracy | F1-Score |
|---|---|---|
| Our model (XGBoost) | 77.5% | 77.0% |
| GPTZero | 48.5% | 58.0% |

**Table (6): Confusion matrix for GPTZero.**

| | | | | |
|---|---|---|---|---|
| | Pure AI | 3 | 56 | 0 |
| Actual | Mixed | 0 | 76 | 0 |
| | Pure Human | 0 | 15 | 18 |
| | | Pure AI | Mixed | Pure Human |
| | | | Predicted | |

**Table (6): Confusion matrix for our proposed model.**

| | | | | |
|---|---|---|---|---|
| | Pure AI | 48 | 18 | 0 |
| Actual | Mixed | 7 | 55 | 5 |
| | Pure Human | 0 | 15 | 52 |
| | | Pure AI | Mixed | Pure Human |
| | | | Predicted | |

### *4.5 Advancing Knowledge on the Pedagogical Use of Digital Technology*

The results of this study have significant implications in advancing the pedagogical use of digital technology, particularly in maintaining academic integrity and improving learning environments [49], [50]. As AI-generated content becomes more prevalent in educational settings, the ability to accurately distinguish between human-written and AI-generated text is critical for ensuring fairness, transparency, and the authenticity of student work [51], [52]. The study demonstrates that traditional machine learning models, such as XGBoost and Random Forest, can effectively classify AI-generated text with high accuracy, which can be applied to educational contexts where verifying the originality of student submissions is vital.

In an academic environment where students increasingly have access to powerful generative AI tools like ChatGPT, this research highlights how automated systems can assist educators in identifying instances where AI is used excessively. By incorporating XAI techniques such as LIME, the study also provides transparency, allowing educators to understand why certain content is flagged as AI-generated. This enhances trust in the technology and helps educators make informed decisions, which fosters a balanced approach to integrating AI in education while maintaining ethical standards. The findings have implications for how educators design assessments and encourage original thought. With reliable AI detection tools, instructors can confidently promote digital tools in the classroom for learning purposes while ensuring that students remain accountable for their work. This ensures that AI-enhanced learning environments still prioritize critical thinking, creativity, and student engagement, without compromising academic integrity [53], [54].

The results of this study contribute to the field of educational technology by providing actionable tools and insights for educators facing the challenges of generative AI. By improving the ability to detect AI-generated content, the study helps safeguard academic standards, ensuring that digital technologies are used to support, rather than undermine, student learning outcomes. This aligns with the broader goals of enhancing the pedagogical use of digital technology to create equitable and effective learning environments [50], [55].

## 5. Conclusion

This study seeks to advance the pedagogical use of digital technology by providing tools to detect AI-generated content in educational settings, which promotes academic integrity and fairness. By leveraging machine learning models including traditional ML, DL, and XAI techniques, the study helps educators identify AI use in student work, ensuring transparency and accountability. These findings support the ethical integration of AI in education, which helps maintain academic standards while fostering digital literacy and critical thinking in learning environments. This study proposes a model that distinguish between human-written and AI-generated text, which has become a critical challenge, particularly in fields like cybersecurity. This study highlights the importance and practical applications of this distinction, not only within cybersecurity field but also in academic writing and business operations. We tested various ML and DL algorithms on a generated dataset that contains cybersecurity articles written by humans and AI-generated articles with the same topic by LLMs (specifically, ChatGPT). We demonstrated the high performance of traditional ML algorithms, specifically XGBoost and RF, to accurately classify AI-generated content with an accuracy of 83% and 81% respectively and with minimal misclassification rates. We also showed in this experiment that classifying relatively smaller content (e.g., paragraphs) is more challenging than classifying larger ones (e.g., articles).

We then used LIME, as an XAI model, to elucidate the discriminative features that influence the XGBoost model's predictions. Results offered insights into the characteristics that differentiate human-written content from AI-generated text on the dataset level and on the instance level. It showed that humans tend to use more practical and action-oriented language related to security (e.g., virus, allow, and use) while LLMs use more of an abstract and formal words such as "realm," "employ," "serve," and "establish,".

The main reveal of the comparative analysis between GPTZero and our proposed model showed that a narrowly focused and fine-tuned AI system can outperform more generalized AI systems like GPTZero in specific tasks. This provides evidence of the effectiveness of tailoring AI models to specific datasets and tasks, where precision and performance can be significantly improved with a more targeted approach. GPTZero model showed an accuracy of 48.5% with about 16% of the cases that were not recognized, while our proposed model achieved about 77.5% accuracy. GPTZero had tendency to classify uncertain cases as either mixed or unrecognized rather than taking the risk of misclassifying them as Pure AI or Pure Human. However, our proposed model showed a more balanced performance across the three classes, namely, Pure AI, Pure Human, and mixed.